\documentclass[runningheads]{llncs}
\usepackage{xcolor}

\usepackage{lineno}
\usepackage[hidelinks]{hyperref}

\usepackage{graphicx}
\usepackage{booktabs}
\usepackage{subfig}
\usepackage{wrapfig}

\usepackage[linesnumbered,ruled,vlined]{algorithm2e}

\SetCommentSty{mycommfont}

\usepackage{dsfont}
\usepackage{amsmath}
\usepackage{amssymb}
\usepackage{mathtools}
\usepackage{bm}
\usepackage{forest}
\usepackage{tikz}
\usetikzlibrary{trees,bayesnet}
\usepackage{cleveref}
\usepackage{tcolorbox}
\usepackage[tight]{units}

\Crefname{equation}{Eq.}{Eqs.}
\Crefname{figure}{Fig.}{Figs.}
\Crefname{tabular}{Tab.}{Tabs.}

\usepackage{amsfonts}       
\usepackage{nicefrac}       
\usepackage{microtype}      

\DeclareMathOperator{\E}{\mathbb{E}}

\DeclareMathOperator{\Dir}{\mathrm{Dir}}
\DeclareMathOperator{\Cat}{\mathrm{Categorical}}
\DeclareMathOperator{\Gem}{\mathrm{GEM}}

\DeclareMathOperator{\DP}{\mathrm{DP}}
\DeclareMathOperator{\U}{\mathrm{Unif}}

\DeclareMathOperator{\Ga}{\mathrm{Ga}}
\DeclareMathOperator{\B}{\mathrm{Beta}}

\DeclareMathOperator{\N}{\mathrm{Normal}}

\DeclareMathOperator{\post}{\mathcal{P}}
\DeclareMathOperator{\ncrp}{\mathrm{nCRP}}
\newcommand{\tree}{\mathcal{H}}
\newcommand{\m}[1]{\mathbf{#1}}
\newcommand{\data}{\mathbf{X}}
\newcommand{\eset}{E_{\tree}}
\newcommand{\nset}{Z_{\tree}}

\begin{document}
\title{Inferring Hierarchical Mixture Structures: \\ A Bayesian Nonparametric Approach}
%
\titlerunning{Bayesian Nonparametric Hierarchical Mixture Clustering}
%
\author{Weipeng Huang\inst{1} \and
Nishma Laitonjam\inst{1} \and
Guangyuan Piao\inst{2} \and
Neil J. Hurley\inst{1}}
\authorrunning{W. Huang et al.}
%
\institute{
Insight Centre for Data Analytics, University College Dublin, Ireland \\
\email{first.last@insight-centre.org}
\and
Department of Computer Science, Maynooth University, Ireland \\
\email{guangyuan.piao@mu.ie}
}

\maketitle              

\begin{abstract}

We present a Bayesian Nonparametric model for Hierarchical Clustering (HC).
Such a model has two main components.
The first component is the random walk process from parent to child in the hierarchy and we apply nested Chinese Restaurant Process (nCRP).
Then, the second part is the diffusion process from parent to child where we employ Hierarchical Dirichlet Process Mixture Model (HDPMM).
This is different from the common choice which is Gaussian-to-Gaussian.
We demonstrate the properties of the model and propose a Markov Chain Monte Carlo procedure with elegantly analytical updating steps for inferring the model variables.
Experiments on the real-world datasets show that our method obtains reasonable hierarchies and remarkable empirical results according to some well known metrics.
\end{abstract}

\section{Introduction}
We study the problem of Hierarchical Clustering (HC) via a Bayesian Nonparametric (BNP) modelling perspective.
A BNP proposes a generative model for the observed data whose dimension is not fixed, but rather is learned from the data. In the case of HC, this allows the structure of the hierarchy to be inferred along with its clusters.
Considering a model for generating the data, a BNP for HC typically associates each node in the hierarchy with a particular choice of parameters.
Then the model consists of two components: 1) the random process for generating a path through the hierarchy from the root node to a leaf; 2) a parent-to-child transition kernel that models how  a child node's parameters are related to those of its parent.
The generative model posits that each observation firstly selects a path randomly, and is then sampled through the distribution associated with the leaf node of the path.
\begin{figure}
\centering
\subfloat[Gaussian node \label{fg:gauss-node}]{
  \includegraphics[scale=0.21]{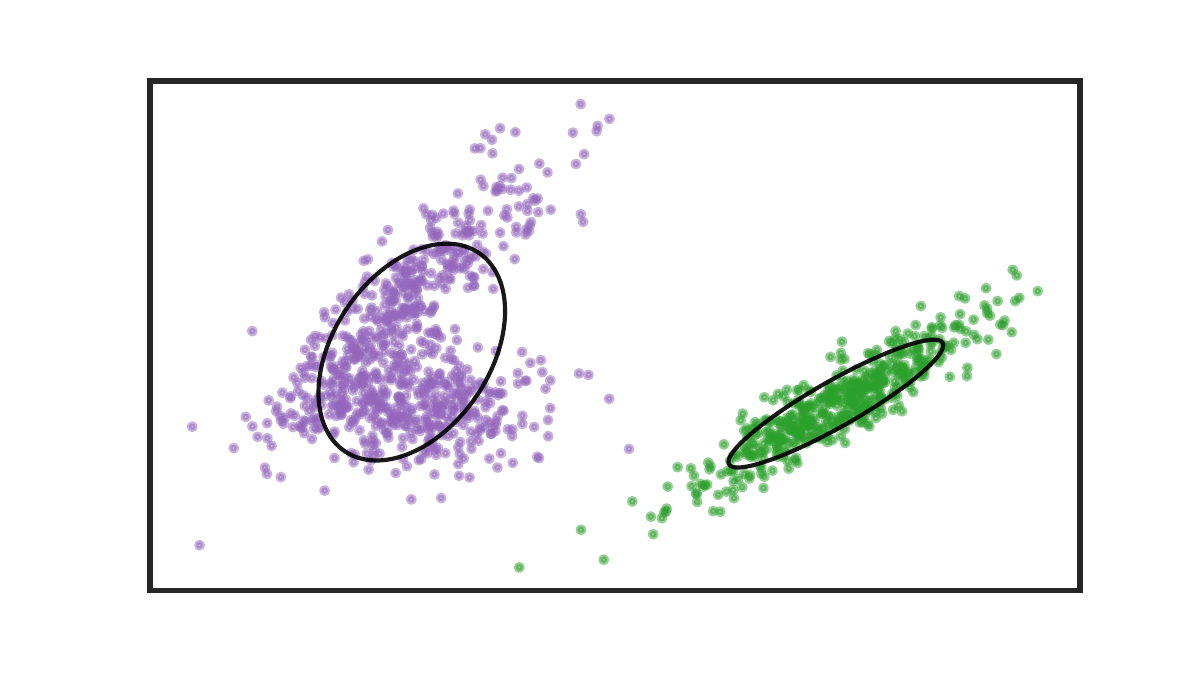}
}
~
\subfloat[Mixture node \label{fg:mixture-node}]{
  \includegraphics[scale=0.21]{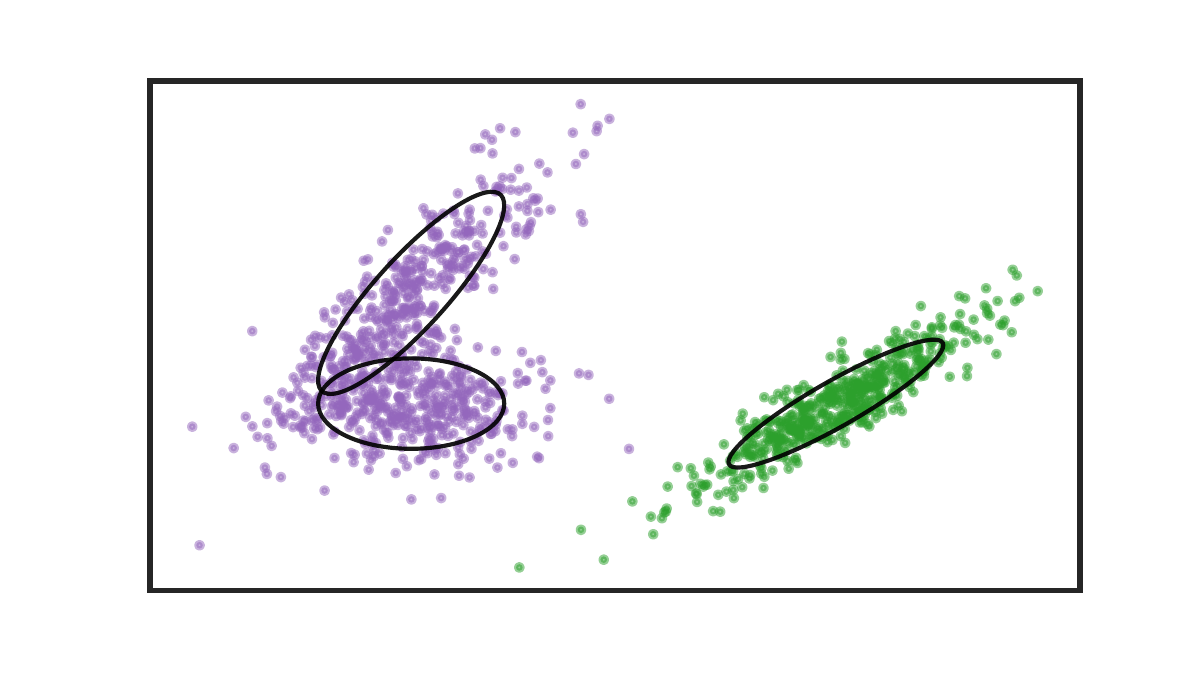}
}
\caption{A simple example of the two nodes (distinguished by colour) at the first layer after the root node.~\Cref{fg:gauss-node} depicts a manner of clustering the data with Gaussian-to-Gaussian node transition from root to the level 1, and~\Cref{fg:mixture-node} is the version of applying HDPMM.}
\label{fg:kernel-comparison}
\end{figure}

An early example of the parametric generative probabilistic model is the Gaussian tree generative process~\cite{williams2000mcmc}, such that each non-root node is sampled from a Gaussian distribution with the mean of the parent node and predefined level-wise covariances.
In general, for the first component, there are some well-known random processes, e.g. the nested Chinese Restaurant Process (nCRP)~\cite{blei2010nested,adams2010tree}, the Dirichlet Diffusion Tree (DTT)~\cite{neal2003density}, the Pitman-Yor Diffusion Tree (PYDT)~\cite{Knowles2015}, which is a generalisation of the DTT, and the  tree-structured stick-breaking construction (TSSB)~\cite{adams2010tree}, which generalises  the (nCRP).
It is proved in~\cite{Knowles2015}, that the nCRP and PYDT (and so also the DTT) are asymptotically equivalent---thus, this work focuses on the second component and selects the nCRP for the first component, given its simplicity.
Existing BNP methods mostly utilise Gaussian-to-Gaussian (G2G) diffusion kernels for the node parameters~\cite{neal2003density,williams2000mcmc,adams2010tree,Knowles2015}.
Within such a setting, e.g., if a parent node's distribution is (denoted by $\sim$) $\N(\bm{\mu}_p, \m{\Sigma}_p)$, its child has the node parameters $\bm{\mu}_c \sim \N(\bm{\mu}_p, \m{\Sigma}_p)$ and is associated with the data distribution $\N(\bm{\mu}_c, \m{\Sigma}_c)$ where $\m{\Sigma}_p$ and $\m{\Sigma}_c$ can be predefined hyperparameters.

G2G kernels are easy to apply but somehow lack the ability to handle more complex patterns.
In our work, we investigate the Hierarchical Dirichlet Process Mixture Model (HDPMM) for the parent-to-node transition.
In this setting, each node is actually a mixture distribution and the node parameters maintain the mixture weights for a global book of components.
Then, the parent-to-node transition follows a Hierarchical Dirichlet Process (HDP)~\cite{teh2010hierarchical}. As a simple example, in~\Cref{fg:kernel-comparison}, a single level of clustering is applied to the shown data points, using G2G~(\Cref{fg:gauss-node}) and HDPMM~(\Cref{fg:mixture-node}). The HDPMM fits a mixture model to the purple nodes, which can be further refined in lower levels of the tree.
Note that, the HDP~\cite{Teh2006hier} is discussed in~\cite{adams2010tree}, for applications based on Latent Dirichlet Allocations.
However, our setting HDPMM is more suited to general clustering.


\subsubsection{Related work}
Our focus is on statistical methods, despite that there are many non-statistical methods for HC e.g. \cite{ward1963hierarchical,Monath2019,kobren2017hierarchical,charikar2019hierarchical}, to name just a few.
Statistical extensions to Agglomerative clustering (AC) have been proposed in~\cite{stolcke1993hidden,iwayama1995hierarchical,heller2005bhc,lee2015bayesian} but these are not generative models.
On the other hand, Teh et al.~\cite{teh2008bayesian} has proposed applying the Kingman's coalescent as a prior to the HC, which is similar in spirit to DDT and PYDT, however it is a backward generative process.

It is worth mentioning that the tree generation and transition kernels that we exploit have been used in a number of other contexts, beyond HC. For example, Paisley et al.~\cite{paisley2015nested} discuss the nested Hierarchical Dirichlet Process (nHDP) for topic modelling which, similar to the method we present here, uses the nCRP to navigate through the hierarchy, but associates nodes in the tree with topics. The nHDP first generates a global topic tree using the nCRP where each node relates to one atom (topic) drawn from the base distribution. In summary, nHDP constitutes multiple trees with one global tree and many local trees, whereas our construction has a single tree capturing the hierarchical structure of the data. Ahmed et al.~\cite{Ahmed2013} also proposed a model that is very similar to the nHDP, but appealed to different inference procedures.

\section{Preliminary}
\paragraph{Dirichlet Process}
We briefly describe the CRP and the stick-breaking process which are two forms of the Dirichlet Process (DP).

In the CRP, we imagine a Chinese restaurant consisting of an infinite number of tables, each with sufficient capacity to seat an infinite number of customers.
A customer enters the restaurant and picks one table  at which to sit.
The $n^{th}$ customer picks a table based on the previous customers' choices.
That is, assuming  $c_n$ is the table assignment label for customer $n$ and $N_k$ is the number of customers at table $k$, one obtains
\begin{align*}
p(c_{n+1} = k \mid \bm{c}_{1:n}) =
\begin{cases}
  \frac {N_k} {n + \alpha} & \mbox{existing }k \\
  \frac {\alpha} {n + \alpha} & \mbox{new }k
\end{cases}
~~
\theta^*_{n+1} \mid \bm{\theta^*}_{1:n} &\sim \frac {\alpha} {n+\alpha} H + \sum_{k=1}^{K} \frac {N_k} {n+\alpha} \delta_{\theta_k }
\end{align*}
where $\bm{c}_{1:n} = \{ c_1, \ldots, c_{n} \}$, likewise for $\bm{\theta^*}_{1:n}$.
The right hand side indicates how the parameter $\theta^*_{n+1}$ is drawn given the previous parameters, where each $\theta_k$ is sampled from a base measure $H$, and $\theta_1, \ldots,\theta_K$ are the unique values among $\theta^*_1,\ldots,\theta^*_{n}$.

Denoting a distribution by $G$, the stick-breaking process can depicted by
$G = \sum_{k=1}^{\infty} \beta_k \delta_{\theta_k}$, $ \{\theta_k\}_{k=1}^\infty \sim H$ and $\bm{\beta} \sim \Gem(\alpha)$.
Also, $\Gem$ (named after Griffiths, Engen and McCloskey) known as a stick-breaking process,
is analogous to iteratively breaking a portion from the remaining stick which has the initial length $1$. In particular, we write $\bm{\beta} \sim \Gem(\alpha)$ when $u_k \sim \B(1, \alpha)$, $\beta_1 = u_1$, and $\beta_k = u_k \prod_{l=1}^{k-1} (1-u_l)$.

\paragraph{Nested CRP}
In the nCRP~\cite{blei2010nested}, customers arrive at a restaurant and choose a table according to the CRP, but at each chosen table, there is a card leading to another restaurant, which the customer visits the next day, again using the CRP.
Each restaurant is associated with only a single card. After $L$ days, the customer has visited $L$ restaurants, by choosing a particular path in an infinitely branching hierarchy of restaurants.

\paragraph{Hierarchical Dirichlet Process Mixture Model}
\label{sec:hdp}
When the number of mixture components is infinite, we connect components along a path in the hierarchy using a HDP.
The 1-level HDP~\cite{Teh2006hier} connects a set of DPs, $G_j$, to a common base DP, $G_0$.
It can be simply written as
$G_j \sim \DP(\gamma, G_0)$ and $G_0 \sim \DP(\gamma_0, H)$.
It has several equivalent representations while we will focus on the following form:
$\bm{\beta}_0 \sim \Gem(\gamma_0)$, $\bm{\beta}_j \sim \DP(\gamma, \bm{\beta}_0)$, and $\theta_k \sim H$
to obtain $G_0 = \sum_{k=1}^\infty \beta_{0 k} \delta_{\theta_k}$ and then $G_j = \sum_k \beta_{jk} \delta_{\theta_{k}}$. Hence $G_j$ has the same components as $G_0$ but with different mixing proportions.
It may be shown that $\bm{\beta}_j$ can be sampled by firstly drawing $u_{jk} \sim \B(\gamma \beta_k, \gamma(1 - \sum_{\ell=1}^k \beta_\ell))$ and then $\beta_{jk} = u_{jk} \prod_{\ell=1}^{k-1} (1 - u_{j\ell})$.
Considering each data in $x$ belongs to one of the mixture models and in particular $\bm{x}_n$ belongs to the group associated with $G_j$, the 1-level HDPMM is completed by having
$\bm{x}_n \sim F(\theta_{c_n})$ where $c_n \sim \Cat(\bm{\beta}_j)$.
This process can be extended to multiple levels, by defining another level of DPs with $G_j$ as a base distribution and so on to higher levels.
In fact, we can build a hierarchy where, for any length $L$ path in the hierarchy, the nodes in the path correspond to an $L$-level HDP.

\section{Generative Process}
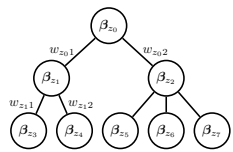
\begin{wrapfigure}{r}{0.4\textwidth}
 \centering
 \scalebox{0.5}{
 \begin{forest}
   for tree={
     circle, draw, very thick, edge={very thick}
   }
   [$\bm{\beta}_{z_0} $
     [$\bm{\beta}_{z_1}$, edge label={node[midway,left]{$w_{z_0 1}$}}
       [$\bm{\beta}_{z_3} $, edge label={node[midway,left]{$w_{z_1 1}$}}]
       [$\bm{\beta}_{z_4} $, edge label={node[midway,right]{$w_{z_1 2}$}}]
     ]
     [$\bm{\beta}_{z_2} $, edge label={node[midway,right]{$w_{z_0 2}$}}
       [$\bm{\beta}_{z_5} $]
       [$\bm{\beta}_{z_6} $]
       [$\bm{\beta}_{z_7} $]
     ]
   ]
 \end{forest}
 }
 \caption{One example of a BHMC hierarchy}
 \label{fg:tree-example}
 \vspace{-10pt}
\end{wrapfigure}

We call our model BHMC, for Bayesian Hierarchical Mixture Clustering and illustrate it using \Cref{fg:tree-example}.

Let $z$ be the label for a certain node in the tree.
We can denote the probability to choose the first child under $z$ by $w_{z 1}$.
Also, let us denote the mixing proportion for a node $z$ by $\bm{\beta}_z$, and denote the global component assignment for the $n^{th}$ data item, $\bm{x}_n$, by $c_n$. To draw $\bm{x}_n$, the hierarchy is traversed to a leaf node, $z$, at which $c_n$ is drawn from mixing proportions $\bm{\beta}_z$.
A path through the hierarchy is denoted by a vector e.g. $\bm{v} = \{z_0, z_1, z_4\}$.
In a finite setting, with a fixed number of branches, such a path would be generated by sampling from weights $\bm{w}_{z_0} \sim \Dir(\alpha)$, at the first level, then $\bm{w}_{z_1} \sim \Dir(\alpha)$ at the second level and so on. The nCRP enables the path to be sampled from infinitely branched nodes.
Mixing proportions, $\bm{\beta}_z$, along a path are connected via a multilevel HDP. Hence, if the probability of a mixture component goes to zero at any node in the tree, it will remain zero for any descendant nodes. Moreover, the smaller $\gamma$ is, the sparser the resulting distribution drawn from the HDP~\cite{murphy2012machine}.

The generative process with an infinite configuration is shown in~\Cref{alg:generative}.
For a finite setting, the mixing proportions are on a finite set and Line~1 should be changed to $\bm{\beta}_{z_0} \sim \Dir(\gamma_0/K, \dots, \gamma_0/K)$.
Correspondingly, Line 9 has to be changed to $\bm{\beta}_{z'} \sim \Dir(\gamma \bm{\beta}_z)$ according to the preliminary section.
%


\subsection{Properties}
Let us denote the tree by $\tree = (\nset, \eset)$.
First, the node set is $\nset=\{z_0, \dots, z_{M-1}\}$ where $M$ is the number of nodes.
Then, $\eset$ is the set of edges in the tree, where $(z,z') \in \eset$ means that there exists some path $\bm{v}_n$, such that $\bm{x}_n$ moves from $z$ to $z'$ in the path $\bm{v}_n$.
Hence, we infer the following variables:
$\m{B} = \{\bm{\beta}_{z_0}, \dots, \bm{\beta}_{z_{M-1}}\}$, the mixing proportions for the components in the node $z$;
$\bm{\theta}=\{\theta_1, \dots, \theta_K\}$, component parameters;
$\m{V} = \{\bm{v}_1, \dots, \bm{v}_N \}$ where each $\bm{v}_n = \{v_{n 0}, v_{n 1}, \ldots, v_{n L}\}$ is the ordered set of nodes in $\nset$ corresponding to the path of $\bm{x}_n$;
$\bm{c} = \{c_1, \dots, c_N\}$, the component label for all the observations.
Denoting $\Phi = \{{\gamma_0}, \gamma, \alpha, H, L\}$, we focus on the marginal prior $p(\m{V}, \bm{\theta}, \m{B} \mid \Phi)$ and obtain $p(\m{V}, \bm{\theta}, \m{B} \mid \Phi) = p(\m{B} \mid \m{V}, \gamma_0, \gamma) p(\m{V} \mid \alpha, L) p(\bm{\theta} \mid H)$.

The first term is $p(\m{B} \mid \m{V}, \gamma_0, \gamma) = p\left(\bm{\beta}_0 \mid {\gamma_0}\right)\prod_{(z, z') \in \eset} p(\bm{\beta}_{z'} \mid \gamma, \bm{\beta}_{z})$.
To expand the second term, we first denote by $m_{z}$ the number of children of $z$.
Hence, the CRP probability for a set of clusters $\{z'_{n}\}_{n=1}^{m_z}$ under the same parent $z$~\cite{ferguson1973bayesian,antoniak1974mixtures}:
\[
p(\{z'_{n}\}_{n=1}^{m_z} \mid \alpha) = \frac {\alpha^{m_z} \Gamma(\alpha)} {\Gamma(N_{z} + \alpha)} \prod_{n=1}^{m_z} \Gamma\left(N_{z'_n}\right)
\]
where $N_z$ is the number of observations in $z$.
With this equation, one can observe the exchangeability of the order of arriving customers---the probability of obtaining such a partition is not dependent on the order.
The tree $\tree$ is constructed via $\m{V}$ with the empty nodes all removed.
For such a tree, the above result can be extended to
\[
\textstyle
p(\m{V} \mid \alpha)
= \Gamma(\alpha)^{\lvert \mathcal{I}_{\tree} \rvert}\prod_{z \in \mathcal{I}_{\tree}} \frac {\alpha^{m_{z}}} {\Gamma(N_{z}+\alpha)} \prod_{(z, z') \in \eset} \Gamma(N_{z'})
\]
where $\mathcal{I}_{\tree}$ is the set of internal nodes in $\tree$.
Next, we obtain $p(\bm{\theta} \mid H) = \prod_{k=1}^K p(\theta_k \mid H)$.
The likelihood for a single observation  is
\begin{equation}
\label{eq:lkhd}
\textstyle p\left(\bm{x}_n \mid v_{n L}, \bm{\theta}, \bm{\beta}_{v_{n L}}\right) =
 \sum_{k=1}^K \beta_{v_{n L} k} f(\bm{x}_n; \theta_k) + \beta_{v_{n L}}^* f^*(\bm{x}_n)
\end{equation}
where $\beta^*_{z}$ denotes the probability of drawing a new component which is always the last element in the vector $\bm{\beta}_z$.
Here, $f(\cdot)$ is the corresponding density function for the distribution $F$ and we obtain $f(\bm{x}; \theta) \equiv p(\bm{x} \mid \theta)$.
Furthermore, $f^*(\bm{x}) = \int p(\bm{x} \mid \theta) d p(\theta \mid H)$.
The above presentation is similar to the Poly\'{a} urn construction of DP~\cite{pitman1996some}, which trims the infinite setting of DP to a finite configuration.
Then, one can see $ p(\data \mid \m{V}, \bm{\theta}, \m{B}) = \prod_{n} p\left(\bm{x}_n \mid v_{n L}, \bm{\theta}, \bm{\beta}_{v_{n L}}\right)$.
Finally, the unnormalised posterior is
$
p(\m{V}, \bm{\theta}, \m{B} \mid \data, \Phi) \propto p(\data \mid \m{V}, \bm{\theta}, \m{B}) p(\m{V}, \bm{\theta}, \m{B} \mid \Phi)
$.

\hspace{-4ex}
\begin{minipage}{.54\textwidth}
\begin{algorithm}[H]
\DontPrintSemicolon
\caption{Generative process (infinite)}\label{alg:generative}
Sample $\bm{\beta}_{z_0} \sim \Gem(\gamma_0)$ \label{GPalgo:infi:root}\\
Sample $\theta_1, \theta_2, \ldots \sim H$ \\
\For{$n=1 \dots N$}{
  $v_{n 0} \gets z_0$ \\
  \For{$\ell = 1 \dots L$} {
    Sample $v_{n \ell}$ using CRP($\alpha$) \\
    $z, z' \gets v_{n (\ell-1)}, v_{n \ell}$ \\
    \If{$z'$ is new}{
      Sample $\bm{\beta}_{z'} \sim \DP\left(\gamma, \bm{\beta}_{z}\right)$ \label{GPalgo:infi:nonroot}\\
      Attach $(z, z')$ to the tree
    }
  }
  Sample $c_n \sim \Cat\left({\beta}_{v_{n L}}\right)$ \\
  Sample $\bm{x}_n \sim F(\theta_{c_n})$ \\
}
\end{algorithm}
\end{minipage}
~
\begin{minipage}{.45\textwidth}
\begin{algorithm}[H]
\DontPrintSemicolon
\caption{MH sampler}\label{alg:scheme-a}
\tcp{$\epsilon$: stopping threshold}
Sample $\bm{\beta}_{z_0}$ until $\beta^*_{z_0} < \epsilon$ \\
\For{$\bm{x}_n \in \textsc{Shuffled}(\data)$}{
  Clean up $c_n$ and $\bm{v}_n$ \\
  Sample $\hat{\bm{v}}_n$ (and possibly new $\bm{\beta}$) through the generative process \\
  $s \sim \U(0, 1)$ \\
  \lIf{$s \le \mathcal{A}$}{
    $\bm{v}_n \gets \hat{\bm{v}}_n$
  }
  Sample ${c}_n$ using a Gibbs step by~\Cref{eq:update-c}
}
Update $\m{B}$ by~\Cref{eq:update-b-root,eq:update-b} \\
Update ${\bm{\theta}}$ by~\Cref{eq:update-phi}
\end{algorithm}
\end{minipage}%

\section{Inference}
\label{sec:inference}
We appeal to Markov Chain Monte Carlo (MCMC) for inferring the model.
One crucial property for facilitating the sampling procedure is exchangeability, which, as noted in the previous section, follows from the model's connection to the CRP.

\paragraph{Sampling $\m{V}$ and $\bm{c}$}
Following~\cite{blei2010nested}, we sample a path for a data index as a complete variable using nCRP and decide to preserve the change based on a Metropolis-Hastings (MH) step.

Recall that the component will only be drawn at the leaf.
Thus,
\begin{align}
\label{eq:update-c}
p \left(c_n = k \mid \bm{x}_n, v_{n L}=z, \m{B}, \bm{\theta}\right)
\propto
\begin{cases}
 \beta_{z k} f(\bm{x}_n; \theta_k) & \mbox{existing } k \\
 \beta^*_z f^*(\bm{x}_n) & \mbox{new } k \, .
\end{cases}
\end{align}
Sampling the set $\bm{c}$ is necessary for updating the set of mixtures at each node i.e., $\m{B}$.
Our MH scheme applies a partially collapsed Gibbs step.
Following the principles in~\cite{Dyk2015MH}, our algorithm first samples $\m{V}$ with $\bm{c}$ being collapsed out, and subsequently updates $\bm{c}$ based on $\m{V}$.

\paragraph{Sampling $\m{B}$}
It is straightforward to decide the sampling for a leaf node.
For non-root nodes, we would like to find out $p\left(c_{n} = k \mid \m{B} \setminus \{ \m{B}_{\bar{z}}\}, v_{n (L-1)} = z \right)$ where $\m{B}_{\bar{z}}$ is the set of mixing proportions in the sub-tree rooted at $z$ excluding $\bm{\beta}_z$.
We write $p\left(c_{n} = k \mid \m{B} \setminus \{ \m{B}_{\bar{z}}\}, v_{n (L-1)} = z \right)$ to be
\[
p\left( v_{n L} = z' \mid v_{n (L-1)} = z \right)p\left(c_n = k \mid \bm{\beta}_{z}, v_{n L}=z', v_{n (L-1)}=z \right) \, .
\]
Logically, $p(c_n = k \mid \m{B}, \bm{v}_n) = p\left(c_n = k \mid \{\bm{\beta}_{v_{n \ell}} \}_{\ell=0}^{L} \right)$
which is then $p\left(c_n = k \mid \bm{\beta}_{v_{n L}}\right)$.
We also derive $p\left(c_n = k \mid \bm{\beta}_{z}, v_{n L}=z', v_{n (L-1)}=z \right)$ to be
\begin{align}
\int p(c_n = k \mid \bm{\beta}_{z'}) d p(\bm{\beta}_{z'} \mid \bm{\beta}_{z})
&= \int p\left(c_n = k \mid \bm{\beta}_{z'} \right) \frac {\Gamma\left(\sum_{j=1}^K \gamma \beta_{z j}\right)} {\prod_{j=1}^K \Gamma(\gamma \beta_{z j})} \prod_{j=1}^K \beta_{z' k}^{\gamma \beta_{z k} - 1} d \bm{\beta}_{z'} \nonumber \\
&= \frac {\Gamma\left(\sum_{j=1}^K \gamma \beta_{z j}\right)} {\prod_{j=1}^K \Gamma(\gamma \beta_{z j})} \int \prod_{j=1}^K \beta_{z' j}^{\mathds{1}\{c_{n} = k\} + \gamma \beta_{z k} - 1} d \bm{\beta}_{z'} \nonumber \\
&= \frac {\Gamma\left(\sum_{j=1}^K \gamma \beta_{z j}\right)} {\prod_{j=1}^K \Gamma(\gamma \beta_{z j})} \frac {\prod_{j=1}^K \Gamma(\mathds{1}\{c_{n} = k\} + \gamma \beta_{z j})} {\Gamma(1 + \sum_{j=1}^K \gamma \beta_{z j})} = \beta_{z k} \nonumber \,
\end{align}
given that $\Gamma(x + 1) = x \Gamma(x)$ holds when $x$ is any complex number except the non-positive integers.
Therefore, $p\left(c_{n} = k \mid \m{B} \setminus \{ \m{B}_{\bar{z}}\}, v_{n (L-1)} = z \right)=\beta_{z k}$.
This indicates that, marginalising out the subtree rooted at $z$, the component assignment is thought to be drawn from $\bm{\beta}_{z}$, which can be seen through induction.
Therefore, it allows us to conduct the size-biased permutation,
\begin{align}
\mbox{root } z_0 & &\beta_{z_0 1}, \dots, \beta_{z_0 K}, \beta_{z_0}^* &\sim \Dir\left(N_{z_0 1}, \dots, N_{z_0 K}, \gamma_0 \right) \label{eq:update-b-root} \\
\forall z': (z, z') \in \eset & & {\beta}_{z'1}, \dots, \beta_{z'K}, \beta^*_{z'} &\sim \Dir\left({\tilde N}_{z'1}, \dots, {\tilde N}_{z'K}, \gamma \beta^*_{z} \right) \label{eq:update-b}
\end{align}
where $\tilde{N}_{z' k} = N_{z'k} + \gamma \beta_{z k}$.
\Cref{eq:update-b-root} employs a Poly\'{a} urn posterior construction of the DP to preserve the exchangeability when carrying out the size-biased permutation~\cite{pitman1996some}.
This step in our inference enables an analytical form of node parameter update.


\paragraph{Sampling $\bm{\theta}$}
Even though there are many options for $H$, we choose a Gaussian distribution in this paper.
With $\theta_k \coloneqq \bm{\mu}_k$,
we define $f(\bm{x}; \bm{\mu}_k) = \N(\bm{x}; \bm{\mu}_k, \m{\Sigma})$ and $H = \N(\bm{\mu}_0, \m{\Sigma}_0)$, where covariance matrices $\m{\Sigma}$ and $\m{\Sigma}_0$ are known and fixed.
Collapsing out the unused terms, we can write
$p(\theta_k \mid \data, \bm{c}) \sim p(\theta_k \mid H) \prod_{n: c_n=k} p(\bm{x}_n \mid \theta_k) \,.$
Given the conjugacy of a Gaussian prior with a Gaussian of known covariance,
by considering $\bar{\bm{x}}_k = \frac 1 {N_k} \sum_{n} \mathds{1}\{c_n = k\} \bm{x}_n$ and $N_K = \sum_n \mathds{1}\{c_n = k\}$,
we obtain $\bm{\mu}_k \mid \data, \bm{c} \sim \N(\tilde{\bm{\mu}}_k, \tilde{\m{\Sigma}}_k)$ where
\begin{align}
  \label{eq:update-phi}
  \tilde{\bm{\mu}}_k = \tilde{\m{\Sigma}}_k (\m{\Sigma}_0^{-1} \bm{\mu}_0 + N_k \m{\Sigma}^{-1} \bar{\bm{x}}_k)
  \qquad
  \tilde{\m{\Sigma}}_k = (\m{\Sigma}_0^{-1} + N_k \m{\Sigma}^{-1})^{-1} \, .
\end{align}

\subsection{Algorithmic Procedure}
In practice, one useful step of the inference is to truncate the infinite setting of $\bm{\beta}_{z_0}$ to a finite setting.
Referring back to~\Cref{eq:lkhd}, one can have a threshold such that the sampling of $\bm{\beta}_{z_0}$ terminates when the remaining length of the stick $\beta_{z_0}^*$ is shorter than that threshold.
Once a new component is initialised, each node will update its $\bm{\beta}$ by one more stick-breaking step.
That is, for the root node, it samples one $u$ from $\B(1, \gamma_0)$, and assigns $\beta_{{z_0} (K+1)} = \beta_{z_0}^* u$ and the remaining stick length $1-\sum_{k=1}^{K+1} \beta_{{z_0} k}$ as a new $\beta_{z_0}^*$.
For a non-root node $z'$ inheriting from node $z$, we apply the results from the preliminary section such that $u$ is sampled from $\B(\gamma \beta_{z (K+1)}, \gamma \beta_{z}^*)$ and then update in the same manner as the root node.



\Cref{alg:scheme-a} depicts the procedure.
The MH scheme samples a proposal path and the corresponding $\bm{\beta}$'s if new nodes are initialised.
MH considers an acceptance variable $\mathcal{A}$ such that $\mathcal{A}
= \min \left\{1, \frac {\post' q(\m{V}, \m{B} \mid \m{V}', \m{B}')} {\post q(\m{V}', \m{B}' \mid \m{V}, \m{B})} \right\}$.
Here, $\post$ and $\post'$ are the posteriors at the current and proposed states, respectively.
Then, $q$ is the proposal for sampling $\m{V}'$ and $\m{B}'$, which in our case is the nCRP and HDP.
At each iteration for a certain data index $n$, $\m{V}$ changes to $\m{V}'$ by replacing $\bm{v}_n$ with $\bm{v}'_n$.
Thus, in this example, $q(\m{V}' \mid \m{V}) = \ncrp(\bm{v}'_n; \m{V} \setminus \{\bm{v}_n\})$ and vice versa.
The term $q(\m{B} \mid \m{V}) / q(\m{B}' \mid \m{V}')$ is cancelled out by the terms $p(\m{B}' \mid \m{V}') / p(\m{B} \mid \m{V})$ in the posterior, as $q(\m{B} \mid \m{V})$ and $p(\m{B} \mid \m{V})$ are identical.
Apart from that, $\bm{\theta}$ will be updated only after all the paths are decided, and hence gain no changes.
Therefore, for a specific $\bm{x}_n$, we have that
\begin{align*}
\mathcal{A} =
\min \left\{1,~ \frac {p(\bm{x}_n \mid \beta_{v'_{n L}}, v'_{n L}, \bm{\theta}) p(\m{V}') } {p(\bm{x}_n \mid \beta_{v_{n L}}, v_{n L}, \bm{\theta}) p(\m{V})} \frac {\ncrp(\bm{v}_n; \m{V}' \setminus \{\bm{v}'_n\})} {\ncrp(\bm{v}'_n; \m{V} \setminus \{\bm{v}_n\})} \right\}
\end{align*}
given that the likelihood for $\data \setminus \{\bm{x}_n\}$ remains unaltered.
After the paths $\m{V}$ for all the observations are sampled, the process updates $\m{B}$ and $\bm{\theta}$ using the manner discussed above.

\subsection{Time Complexity}
Contrary to first impression, at each iteration, the algorithm complexity is only $N^2$ for the worst case, and log-linear for the expected case.

Assume that the maximum number of children is $M_{z_\ell}^*$ at the level $\ell$, then the cost of sampling a path for a single observation with the nCRP is $O (\sum_{\ell=1}^L M^*_{z_\ell} )$.
After that, sampling $c_n$ is carried out with time $O(K g(D))$ where $g(D)$ is the time for computing the Gaussian likelihood of $D$-dimensional data.
In regard to the global variables, $\m{B}$ will be updated with every node and thus achieves $O(K \lvert \nset \rvert)$.
Lastly, $\bm{\theta}$ will be updated with time $O(K g_s(D))$ where $g_s(D)$ is the complexity of sampling the Gaussian mean.
In addition, we notice $\sum_{\ell=1}^L M^*_{z_\ell} = O(\lvert \nset \rvert)$ and $g(D) = O(g_s(D))$.
Overall, for one iteration, we can summarise the complexity by
$
\textstyle  N \sum_{\ell=1}^L M^*_{z_\ell} + K \lvert \nset \rvert + N K g_s(D) = O((N + K) \lvert \nset \rvert + N K g_s(D)) \, .
$
Since the number of components will not exceed the data size, $K \le N$ holds.
With respect to $\lvert \nset \rvert$, at each level, it is no more than $N$.
The extreme case is that each datum is a node and extends to $L$ levels.
It follows that $\lvert \nset \rvert \le NL$.
Hence, $O((N + K) \lvert \nset \rvert + N K g_s(D)) = O \left(N^2 (L + g_s(D)) \right)$.

The expected number of DP components, considering $\gamma_0$, is $\gamma_0 \log N$ for sufficiently large $N$~\cite{murphy2012machine}. Likewise for the first level in the nCRP, which is then $\alpha \log N$.
Let $\E[\cdot]$ denote the expectation over all random $\tree$ drawn from the BHMC. 
If $N$ is sufficiently large, we can have $\sum_{\ell=1}^L \E [M_{z_\ell}] = O(L\alpha \log N)$, 
as the expected number of nodes will be no greater than $\alpha \log N$.
However, $\E [\lvert \nset \rvert]$ is hard to decide but is known to be $\le \min(NL, (\alpha \log N)^L)$.
If $N \gg 0$ and $\alpha > 0$, then $NL \le (\alpha \log N)^L$.
We can obtain $N (\sum_{\ell=1}^L \E [M_z] + \E [K] g_s(D)) = O(N \log N (\alpha L + \gamma_0 g_s(D)))$ and $\E [K \lvert \nset \rvert] = O(\gamma_0 \log N \times N L)$.
Combining the two expressions, we derive the upper bound $O(N \log N (L + g_s(D)))$ for the average case.

%
\section{Experiments}
For qualitative analysis, we use the datasets: \texttt{Animals}~\cite{kemp2008discovery} and \texttt{MNIST-fashion}~\cite{xiao2017fmnist}.
Following this, we carry out a quantitative analysis for the \texttt{Amazon} text data~\cite{He2016ups} since it contains the cluster labels of all the items at multiple levels.

We fix the parameters for $H$ as
$\bm{\mu}_0 = \m{0}$ and $\m{\Sigma}_0 = \m{I}$.
Additionally, we set the covariance matrix for $F$ as $\m{\Sigma} = \sigma^2 \m{I}$.
We set the number of levels $L$ intuitively based on the data.
For more complex cases, one can consider the theoretical results of the effective length of the nCRP~\cite{Steinhardt2012}.

%

\subsection{Convergence Analysis}
We examine the convergence of the algorithm using \texttt{Animals}.
\Cref{fg:conv} shows the unnormalised log likelihood for five individual simulations.
The simulations quickly reach a certain satisfactory level.
Apart from that, the fluctuation shows that the algorithm keeps searching the solution space over the iterations.

\subsection{Results}
\paragraph{\texttt{Animals}}
This dataset contains 102 binary features, e.g. ``has 6 legs'', ``lives in water'', ``bad temper'', etc.
Observing the heat-map of the empirical covariance of the data, there are not many influential features.
Hence, we employ Principal Component Analysis (PCA)~\cite{murphy2012machine} to reduce it to a seven-dimensional feature space.
The MCMC burns $500$ runs and then reports the one with the greatest complete data likelihood $p(\data, \bm{c}, \m{V} \mid \m{B}, \bm{\theta})$ amongst the following $5,000$ draws (following~\cite{adams2010tree}).

\begin{figure}
\begin{minipage}{0.44\textwidth}
\centering
\includegraphics[scale=.45]{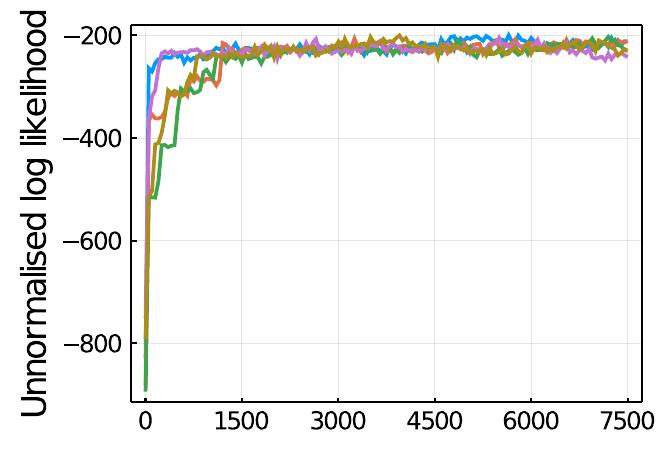}
\caption{Convergence analysis}
\label{fg:conv}
\end{minipage}
~~
\begin{minipage}{0.5\textwidth}
\centering
\includegraphics[scale=5.7]{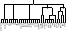}
\caption{The tree of \texttt{Animals}}
\label{fg:animal}
\end{minipage}
\end{figure}

We apply the hyperparameters: $\alpha=0.4, \gamma_0=1, \gamma=0.5, \sigma^{2}=1, L=4$.
\Cref{fg:animal} shows a rather intuitive hierarchical structure.
From the left to the right, there are insects, (potentially) aggressive mammals, herbivores, water animals, and birds.

\paragraph{\texttt{MNIST-fashion}}
This data is a collection of fashion images.
Each image is represented as a $28 \times 28$ vector of grayscale pixel values.
For better visualisation, we sample 100 samples evenly from each class.
PCA transforms the data to 22 dimensions via the asymptotic root mean square optimal threshold~\cite{Gavish2014} for keeping the singular values.
Using the same criterion as for Animal, we output two hierarchies with two sets of hyperparameters.
We set $\alpha=0.5, \gamma_0=1.5, \gamma=2, \sigma^{2}=1, L=5$.
This follows precisely the same running settings as for \texttt{Animals}.

\begin{figure*}
\centering
\includegraphics[scale=2]{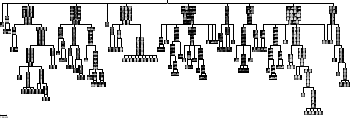}
\caption{The tree of \texttt{MNIST-fashion}}
\label{fg:fmnist}
\end{figure*}

\Cref{fg:fmnist} reflects a property of the CRP which is ``the rich get richer''.
As it is grayscale data, in addition to the shape of the items, other factors affecting the clustering might be, e.g., the foreground/background colour area, the percentage of non-black colours in the image, the darkness/lightness of the item, etc.
The hierarchical structure forms a hierarchy with high purity per level.
Some mis-labelled items are expected in a clustering task.

\paragraph{\texttt{Amazon}}
We uniformly down-sample the indices in the fashion category of \texttt{Amazon} and reserve $2,303$ entries from the data, which contain textual information about items such as their titles and descriptions\footnote{\url{http://jmcauley.ucsd.edu/data/amazon}: the data is available upon request.}.
The data is preprocessed via the method in~\cite{Gavish2014}  and only $190$ features are kept, formed as the Term Frequency and Inverse Document Frequency (TF-IDF) of the items' titles and descriptions.

In this study, we adopt a Bayesian approach to dealing with the hyperparameters.
Due to a compromise to the runtime efficiency, we run BHMC once with $500$ runs for burn-in, during which a MH-based hyperparameter sampling procedure is performed with hyperpriors $\alpha \sim \Ga(3, 1), \gamma_0 \sim \Ga(3.5, 1), \gamma \sim \U(0, 1.5)$ and $\sigma^2 \sim \U(0, 0.1)$.
We limit $\sigma^2$ as mentioned since the feature values in the data are all less than $1$ and some are far less.
In addition, $L$ is fixed to be $7$ (presuming that we have little information about the real number of levels).
The tree with the maximal complete likelihood in the subsequent $2,500$ draws is reported.

We use the evaluation methodology in~\cite{Kuang2013} to compare the clustering results against the ground-truth labels level by level.
We compare 6 levels, which is the maximum branch length of the items in the ground truth.
When extracting the labels from the trees (either for the algorithm outputs or the ground truth), items that are on a path of length less than 6 are extended to level 6 by assigning the same cluster label as that in their last level to the remaining levels.
This is to keep the consistency of the number of items for computing metrics at each level.

For the comparison, we first consider the gold standard AC with Ward distance~\cite{ward1963hierarchical}.
We adopt the existing implementations for PERCH and PYDT\footnote{\url{https://github.com/iesl/xcluster} and \url{https://github.com/davidaknowles/pydt}}.
For PYDT which is sensitive to the hyperparameters, we applied the authors' implemented hyperparameter optimisation to gather the hyperparameters prior to running the repeated simulations.
At each level, we consider four different evaluation metrics, namely, the purity, the normalised mutual information (NMI), the adjusted rand index (ARI), and the F-Measure~\cite{karypis2000comparison}.
Level 1 is the level for the root node.
\begin{figure}[!ht]
\centering
\includegraphics[scale=.32]{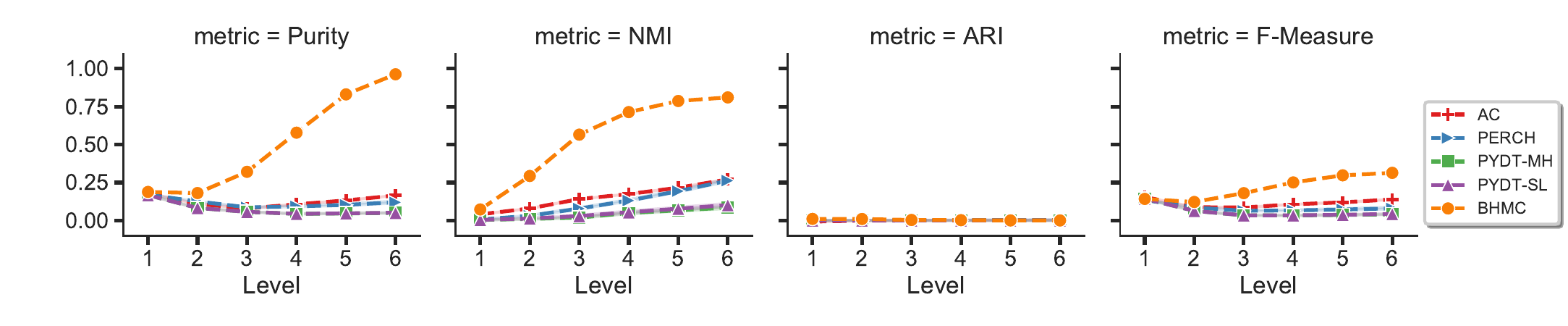}
\caption{Metrics on \texttt{Amazon} by levels}\label{fg:amazon-results}
\end{figure}

\Cref{fg:amazon-results} depicts that our method achieves clearly better scores with respect to purity and NMI.
In the figure, PYDT-SL and PYDT-MH correspond to the slice and the MH sampling solutions, respectively.
As the tree approaches to a lower level, our method also achieves a better performance in F-measure.
For ARI, despite that PERCH performs the best, all numerical values are exceedingly close to $0$.
However, some theoretical work of~\cite{romano2016adjusting} suggests that ARI is more preferred in the scenario that the data contains big and equal-sized clusters.
This is opposed to our ground truth which is highly unbalanced among the clusters at each level.
BHMC does show the potential to perform well according to certain traditional metrics.

\section{Conclusion}
This paper has discussed a new perspective for Bayesian nonparametric HC.
Our model, BHMC, develops an infinitely branching hierarchy of mixture parameters, that are linked along paths in the hierarchy through a multilevel HDP.
A nested CRP is used to select a path in the hierarchy and mixture components are drawn from the mixture distribution in the leaf node of the selected path.
The evaluation shows that BHMC is able to provide good hierarchical clustering results on three real-world datasets with different types of characteristics (i.e., binary, visual and textual) and clearly performs better than other methods with respect to purity and NMI on the Amazon dataset with ground truth, which shows the promising potential of the model.


\subsubsection*{Acknowledgements}
We thank the reviewers for the helpful feedback.
This research has been supported by SFI under the grant SFI/12/RC/2289\_P2.

\bibliographystyle{splncs04}
\bibliography{bhmc}

\begin{thebibliography}{10}
\providecommand{\url}[1]{\texttt{#1}}
\providecommand{\urlprefix}{URL }
\providecommand{\doi}[1]{https://doi.org/#1}

\bibitem{adams2010tree}
Adams, R.P., Ghahramani, Z., Jordan, M.I.: {Tree-Structured Stick Breaking for
  Hierarchical Data}. In: {NeurIPS}, pp. 19--27 (2010)

\bibitem{Ahmed2013}
Ahmed, A., Hong, L., Smola, A.J.: {Nested Chinese Restaurant Franchise
  Processes: Applications to user tracking and document modeling}. ICML
  \textbf{28},  2476--2484 (2013)

\bibitem{antoniak1974mixtures}
Antoniak, C.E.: {Mixtures of Dirichlet Processes with Applications to Bayesian
  Nonparametric Problems}. {The Annals of Statistics} pp. 1152--1174 (1974)

\bibitem{blei2010nested}
Blei, D.M., Griffiths, T.L., Jordan, M.I.: {The Nested Chinese Restaurant
  Process and Bayesian Nonparametric Inference of Topic Hierarchies}. {JACM}
  (2010)

\bibitem{charikar2019hierarchical}
Charikar, M., Chatziafratis, V., Niazadeh, R.: Hierarchical clustering better
  than average-linkage. In: SODA. pp. 2291--2304. SIAM (2019)

\bibitem{williams2000mcmc}
{Christopher K. I. Williams}: {A MCMC Approach to Hierarchical Mixture
  Modelling}. In: Advances in Neural Information Processing Systems. vol.~12,
  pp. 680--686 (2000)

\bibitem{Dyk2015MH}
Dyk, D.A.V., Jiao, X.: {Metropolis-Hastings Within Partially Collapsed Gibbs
  Samplers}. Journal of Computational and Graphical Statistics  \textbf{24}(2),
   301--327 (2015)

\bibitem{ferguson1973bayesian}
Ferguson, T.S.: {A Bayesian Analysis of Some Nonparametric Problems}. {The
  Annals of Statistics} pp. 209--230 (1973)

\bibitem{Gavish2014}
Gavish, M., Donoho, D.L.: {The Optimal Hard Threshold for Singular Values is
  4/$\sqrt{3}$}. IEEE Transactions on Information Theory  \textbf{60}(8),
  5040--5053 (2014)

\bibitem{He2016ups}
He, R., McAuley, J.: {Ups and Downs: Modeling the Visual Evolution of Fashion
  Trends with One-Class Collaborative Filtering}. pp. 507--517. WWW'16 (2016)

\bibitem{heller2005bhc}
Heller, K.A., Ghahramani, Z.: {Bayesian Hierarchical Clustering}. In:
  Proceedings of the 22nd international conference on Machine learning. pp.
  297--304 (2005)

\bibitem{iwayama1995hierarchical}
Iwayama, M., Tokunaga, T.: {Hierarchical Bayesian clustering for automatic text
  classification}. In: IJCAI. vol.~2, pp. 1322--1327 (1995)

\bibitem{karypis2000comparison}
Karypis, M.S.G., Kumar, V.: {A Comparison of Document Clustering Techniques}.
  In: KDD Workshop on Text Mining (2000)

\bibitem{kemp2008discovery}
Kemp, C., Tenenbaum, J.B.: The discovery of structural form. Proceedings of the
  National Academy of Sciences  \textbf{105}(31),  10687--10692 (2008)

\bibitem{Knowles2015}
Knowles, D.A., Ghahramani, Z.: {Pitman Yor Diffusion Trees for Bayesian
  Hierarchical Clustering}. {IEEE TPAMI}  \textbf{37}(2),  271--289 (Feb 2015)

\bibitem{kobren2017hierarchical}
Kobren, A., Monath, N., Krishnamurthy, A., McCallum, A.: {A Hierarchical
  Algorithm for Extreme Clustering}. In: SIGKDD. pp. 255--264. ACM (2017)

\bibitem{Kuang2013}
Kuang, D., Park, H.: {Fast Rank-2 Nonnegative Matrix Factorization for
  Hierarchical Document Clustering}. In: SIGKDD. pp. 739--747. ACM (2013)

\bibitem{lee2015bayesian}
Lee, J., Choi, S.: {Bayesian Hierarchical Clustering with Exponential Family:
  Small-variance Asymptotics and Reducibility}. In: AISTATS. pp. 581--589
  (2015)

\bibitem{Monath2019}
Monath, N., Zaheer, M., Silva, D., McCallum, A., Ahmed, A.: {Gradient-based
  Hierarchical Clustering using Continuous Representations of Trees in
  Hyperbolic Space} pp. 714--722 (2019). \doi{10.1145/3292500.3330997}

\bibitem{murphy2012machine}
Murphy, K.P.: {Machine Learning: A Probabilistic Perspective}. MIT press (2012)

\bibitem{neal2003density}
Neal, R.M.: {Density Modeling and Clustering using Dirichlet Diffusion Trees}.
  {Bayesian Statistics}  \textbf{7},  619--629 (2003)

\bibitem{paisley2015nested}
Paisley, J., Wang, C., Blei, D.M., Jordan, M.I.: {Nested Hierarchical Dirichlet
  Processes}. IEEE. TPAMI  \textbf{37}(2),  256--270 (2015)

\bibitem{pitman1996some}
Pitman, J.: {Some Developments of the Blackwell-MacQueen Urn Scheme}. Lecture
  Notes-Monograph Series pp. 245--267 (1996)

\bibitem{romano2016adjusting}
Romano, S., Vinh, N.X., Bailey, J., Verspoor, K.: {Adjusting for Chance
  Clustering Comparison Measures}. JMLR  \textbf{17}(1),  4635--4666 (2016)

\bibitem{Steinhardt2012}
Steinhardt, J., Ghahramani, Z.: {Flexible Martingale Priors for Deep
  Hierarchies}. International Conference on Artificial Intelligence and
  Statistics (AISTATS)  (2012)

\bibitem{stolcke1993hidden}
Stolcke, A., Omohundro, S.: {Hidden Markov Model Induction by Bayesian Model
  Merging}. In: Advances in neural information processing systems. pp. 11--18
  (1993)

\bibitem{teh2008bayesian}
Teh, Y.W., Daume~III, H., Roy, D.M.: {Bayesian Agglomerative Clustering with
  Coalescents}. In: {NeurIPS}. pp. 1473--1480 (2008)

\bibitem{teh2010hierarchical}
Teh, Y.W., Jordan, M.I.: {Hierarchical Bayesian Nonparametric Models with
  Applications}. {Bayesian Nonparametrics}  \textbf{1},  158--207 (2010)

\bibitem{Teh2006hier}
Teh, Y.W., Jordan, M.I., Beal, M.J., Blei, D.M.: {Hierarchical Dirichlet
  Processes}. {Journal of the American Statistical Association}
  \textbf{101}(476),  1566--1581 (2006)

\bibitem{ward1963hierarchical}
Ward~Jr., J.H.: {Hierarchical Grouping to Optimize an Objective Function}.
  {Journal of the American Statistical Association}  \textbf{58}(301),
  236--244 (1963)

\bibitem{xiao2017fmnist}
Xiao, H., Rasul, K., Vollgraf, R.: Fashion-mnist: a novel image dataset for
  benchmarking machine learning algorithms (2017)

\end{thebibliography}

\end{document}


\twocolumn[
\icmltitle{Bayesian Hierarchical Mixture Clustering using Multilevel Hierarchical Dirichlet Processes: Supplemental Material}



\icmlsetsymbol{equal}{*}

\begin{icmlauthorlist}
\end{icmlauthorlist}

\icmlaffiliation{ucd}{Insight Centre for Data Analytics, University College Dublin, Dublin, Ireland}
\icmlaffiliation{nui}{Insight Centre for Data Analytics, NUI Galway, Dublin, Ireland}

\icmlcorrespondingauthor{Weipeng Huang}{weipeng.huang@insight-centre.org}

\icmlkeywords{Hierarchical Clustering, Clustering, Bayesian Modeling, Bayesian Nonparametric}

\vskip 0.3in
]



\printAffiliationsAndNotice{\icmlEqualContribution} 
\section{Metaphor}
\label{sec:chrf}
Our model can be expressed as a metaphor which is a variant of the CRP.

There is a very large Chinese restaurant franchise.
A customer selects the Chinese restaurant $\mathcal{R}$ first and then the section $\mathcal{S}$ according to nCRP.
We write $v = \{\mathcal{F}, \mathcal{R}, \mathcal{S}\}$.
In a specific section, the customer selects a table $t$ according to CRP($\gamma$).
At each table, one cuisine $c$ will be decided by the first customer sitting at this table. Picking the cuisine is based on CRP($\gamma$).
While at the mean time, a cuisine $c$ will specify on a dish $d$ that day.
The dish is globally maintained and is distributed by $\Dir(\gamma_0/K, \dots, \gamma_0/K)$.
Let $G_0$, $G_1$ and $G_2$ correspond to the distributions in $\mathcal{F}$, $\mathcal{R}$ and $\mathcal{S}$ respectively.
This is equivalent to
$G_2 \sim \DP(\gamma, G_1)$, $G_1 \sim \DP(\gamma, G_0)$, and $G_0 \sim \DP(\gamma, H)$.

\begin{figure}[!ht]
\centering
\tiny
\begin{forest}
  for tree={
    circle, draw
  }
  [$\bm{\beta}_{z_0} $
    [$\bm{\beta}_{z_1}$, edge label={node[midway,left]{$w_{z_0 1}$}}
      [$\bm{\beta}_{z_3} $, edge label={node[midway,left]{$w_{z_1 1}$}}]
      [$\bm{\beta}_{z_4} $, edge label={node[midway,right]{$w_{z_1 2}$}}]
    ]
    [$\bm{\beta}_{z_2} $, edge label={node[midway,right]{$w_{z_0 2}$}}
      [$\bm{\beta}_{z_5} $]
      [$\bm{\beta}_{z_6} $]
      [$\bm{\beta}_{z_7} $]
    ]
  ]
\end{forest}
\caption{One example of CHRF}
\label{fg:tree-example}
\end{figure}


Let us add a few auxiliary variables to explain the connections between CHRF and HDP.
Our metaphor can be represented by
\begin{align}
  &&\bm{v} = \{\mathcal{F}, \mathcal{R}, \mathcal{S}\} &\sim \ncrp(\alpha) \nonumber \\
  \bm{\kappa} &\sim \Dir(\gamma_0 / K, \dots \gamma_0 / K) & k_q(\mathcal{F}) &\sim \D(\bm{\kappa}) \nonumber \\
  \bm{\eta} &\sim \Dir(\gamma / Q, \dots, \gamma / Q) & q_t(\mathcal{R}) &\sim \D(\bm{\eta}) \nonumber \\
  \bm{\tau} &\sim \Dir(\gamma / T, \dots \gamma / T) & t_n(\mathcal{S}) &\sim \D(\bm{\tau}) \nonumber \\
  \phi_k &\sim H & x_n &\sim F\left(\phi_{k_{q_{t_n}}}\right) \label{eq:appendix-gen}
\end{align}
where $\bm{\kappa}$ is the distribution of $K$ dishes, $\bm{\eta}$ is the distribution of $Q$ cuisines, and $\bm{\tau}$ is the distribution of $T$ tables.
The last line omits $\mathcal{F}, \mathcal{R}, \mathcal{S}$ in the notation by assuming the indices $k$, $q$, and $t$ are all globally unique, i.e. one can identify $\mathcal{R}$ via $q$, etc.
We denote
\begin{enumerate}
\item
the table that customer $n$ chooses  in section $\mathcal{S}$ by $t_n(\mathcal{S})$;

\item
the cuisine that table $t$ chooses in restaurant $\mathcal{R}$  by $q_t(\mathcal{R})$;
\begin{itemize}
\item
the cuisine is shared by the customers sitting in table $t$ that selects cuisine $q_t$ in restaurant $\mathcal{R}$
\end{itemize}

\item
the dish that cuisine $q$ chooses  in franchise $\mathcal{F}$  by $k_q(\mathcal{F})$;
\begin{itemize}
\item
the dish is shared by all customers who sit in a table that chooses cuisine $q$, such that  $q$ chooses dish $k$ within the franchise.
\end{itemize}
\end{enumerate}
Given any path $\bm{v}$, the above equations form a Chinese Restaurant Franchise  (CRF) which is the typical representation of the HDP, described in~\cite{sudderth2006graphical,Teh2006hier}.

It is insightful to present the generative process using an equivalent representation of the HDP.
Let us map $\mathcal{F}, \mathcal{R}, \mathcal{S}$ to $z_0$, $z_1$, $z_2$.
As $K \to \infty$,
\[
  G_{z_0} = \sum_{k=1}^K \kappa_k \delta(\phi_k) \sim \DP(\gamma_0, H)
\]
where $\delta(\cdot)$ is the Dirac-delta function. Write $\kappa = \beta_{{z_0}}$ for the mixing proportions of the above components at node $z_0$.

For the node $z_1$, we obtain
$
G_{z_1} = \sum_{k=1}^K \sum_{q: k_q = k} \eta_q \delta(\phi_k) \equiv \sum_{k=1}^K \beta_{{z_1} k} \delta(\phi_k)
$
which follows~\cite[Chapter~2.5.4]{sudderth2006graphical}.
This can be generalized to $z_2$ as well. It implies that the the components $\phi$ are the same for $G_{z_0}$, $G_{z_1}$, and $G_{z_2}$, while the mixing proportions $\beta_{z_0}$, $\beta_{z_1}$ and $\beta_{z_2}$ are distinct.

Teh et al. \cite{Teh2006hier} show that we can obtain $G_{z_0} \sim \DP(\gamma_0, H)$, $G_{z_1} \sim \DP(\gamma, G_{z_0})$, and $G_{z_2} \sim \DP(\gamma, G_{z_1})$ when $K \to \infty, Q \to \infty, T \to \infty$ and that in this infinite setting~\cite{Teh2006hier,sudderth2006graphical}:
\begin{align*}
  \beta_{z_0} \sim \Gem(\gamma_0) \qquad \beta_{z_1} \sim \DP(\gamma, \beta_{z_0}) \qquad
  \beta_{z_2} &\sim \DP(\gamma, \beta_{z_1})\,.
\end{align*}
This main property supports our formulation in the generative process (Line 9 in Algorithm 1).

\section{Derivation}
The notations used here are consistent with the notations in the main paper.
We show the full derivations of Equation (7) in the main paper here.
\begin{align*}
&p\left(c_n = k \pvert \bm{\beta}_{z}, v^L_n=z', v_n^{L-1}=z, \gamma \right) \\
&= \int p\left(c_n = k \pvert \bm{\beta}_{z'}\right) p\left(\bm{\beta}_{z'} \pvert \gamma, \bm{\beta}_{z} \right) d \bm{\beta}_{z'} \\
&= \int p\left(c_i = k \pvert \bm{\beta}_{z'} \right) \frac {\Gamma(\sum_{k=1}^K \gamma \beta_{z k})} {\prod_{k=1}^K \Gamma(\gamma \beta_{z k})} \prod_{k=1}^K \beta_{z' k}^{\gamma \beta_{z k} - 1} d \bm{\beta}_{z'} \\
&= \frac {\Gamma\left(\sum_{j=1}^K \gamma \beta_{z j}\right)} {\prod_{j=1}^K \Gamma(\gamma \beta_{z j})} \int \prod_{j=1}^K \beta_{z' j}^{\mathds{1}\left[c_{n} = k\right] + \gamma \beta_{z k} - 1} d \bm{\beta}_{z'} \\
&= \frac {\Gamma\left(\sum_{j=1}^K \gamma \beta_{z j}\right)} {\prod_{j=1}^K \Gamma\left(\gamma \beta_{z j}\right)} \frac {\prod_{j=1}^K \Gamma\left(\mathds{1}[c_{n} = k] + \gamma \beta_{z j}\right)} {\Gamma\left(1 + \sum_{j=1}^K \gamma \beta_{z j}\right)} \\
&= \frac {\gamma \beta_{z k}} {\sum_{j=1}^K \gamma \beta_{z j} } \\
&= \beta_{z k} \, ,
\end{align*}
given that $\Gamma(x + 1) = x \Gamma(x)$ holds when $x$ is any complex number except the non-positive integers.

\section{Hyperparameters}
As discussed in the full paper, we choose $H$ to be Normal-inverse-Wishart and $F$ to be normally distributed.
Further, denoting the inverse Wishart with $\mathcal{W}^{-1}$, $\NIW(\nu, \lambda,  \mu_0, \Sigma_0)$ samples $\phi = (\mu, \Sigma)$ by
\begin{align*}
  \mu \sim \N\left(\mu_0, \frac 1 {\lambda} \Sigma\right) \qquad
  \Sigma \sim \mathcal{W}^{-1}(\Sigma_0, \nu) \, .
\end{align*}
We fix $\mu_0$ and $\Sigma_0$ to be the empirical mean and covariance of the data respectively.


\paragraph{Animal}
For this small dataset, we observe and decide to set $\alpha=.3, \gamma_0=1, \gamma=1.15, L=4, \lambda=.1, \nu=10$.

\paragraph{MNIST fashion}
We set $\alpha=.5, \gamma_0=5, \gamma=.05, L=4, \lambda=.01, \nu=200$ for Figure 3(a).
Then, let $\alpha=.35, \gamma_0=5, \gamma=1.5, L=4, \lambda=.02, \nu=200$ for Figure 3(b).

\paragraph{Amazon}
Unlike the solutions for the small datasets, Amazon data needs a more informed manner for learning the hyperparameters.
For learning the hyperparameters automatically, we have to specify a hyperpriors for the hyperparameters.

Let us denote the gamma distribution by $\Ga(\cdot)$.
We specify the hyperpriors for Amazon dataset as follows
\begin{align*}
\alpha & \sim \Ga(2, 1) \quad \gamma_0 \sim \Ga(5, 1) \quad \gamma \sim \U(0, 1) \\
L & = \ceil{l} \quad s.t.\quad l \sim \Ga(4, 2) + 1 \\
\nu &\sim 1 / \U(0, 1)+d \quad \lambda \sim \B(5, 5)\,
\end{align*}
where $d$ is the dimension of the data since $\nu > d - 1$ is required to be satisfied.
The uniform distribution can also be replaced with a Beta distribution which can enable probability bias towards the values within some interval.
We choose uniform as it may help random search to look for good values easier.
Using a gamma distribution instead of the inverse of the uniform is also commonly considered.
The distribution for the maximum levels $L$ is chosen to entail that the number around $7$ or $8$ appear the most frequently.

We only repeat the search for $2,500$ rounds and 150 burn-in times, on a subset ($15\%$) of the tested data.
Finally, for Amazon data, we have the hyperparameters as follows:
\begin{align*}
\alpha &= 1.65 \quad \gamma_0 = 8.755 \quad \gamma = 0.5 \\
L &= 8 \\
\lambda &= 0.0373 \quad \nu=209.49 \, .
\end{align*}
No doubt, with more rounds for searching in the hyperparameter space, it may achieve even better parameter set.
Furthermore, setting other hyperpriors may also possibly lead to better performance for the model.
However, there is a trade-off between the runtime efficiency and the performance.

%
%

\bibliography{bhmc}
\bibliographystyle{icml2020}